%% file: main.tex
\documentclass[runningheads]{llncs}

\usepackage{eccv}

\usepackage{eccvabbrv}

\usepackage{graphicx}
\usepackage{booktabs}
\usepackage{diagbox}
\usepackage{booktabs}
\usepackage[marginal]{footmisc} 

\usepackage[accsupp]{axessibility}  

\usepackage[pagebackref,breaklinks,colorlinks,citecolor=eccvblue]{hyperref}

\usepackage{orcidlink}

\newcommand{\yty}[1]{\textcolor{black}{#1}}

\begin{document}

\title{Compress3D: a Compressed Latent Space for 3D Generation from a Single Image} 

\titlerunning{Compress3D}

\author{
Bowen Zhang\inst{1}\thanks{Work done during the internship at IDEA.} \and
Tianyu Yang\inst{2\dag}
Yu Li\inst{2} \and
Lei Zhang\inst{2} \and
Xi Zhao\inst{1}\thanks{Corresponding authors.}
}

\authorrunning{Zhang et al.}

\institute{Xi’an Jiaotong University \and
International Digital Economy Academy (IDEA) \\
}

\maketitle

\input{figures/teaser}

\input{aticle_architectures/abstract}

\input{aticle_architectures/introduction}

\input{aticle_architectures/related_work}

\input{aticle_architectures/method}

\input{aticle_architectures/experiments}

\input{aticle_architectures/conclusion}


%
%
\bibliographystyle{splncs04}
\bibliography{main}
\end{document}

%% file: figures/teaser.tex
\begin{figure}[thbp]
    \centering
    \includegraphics[width=\textwidth]{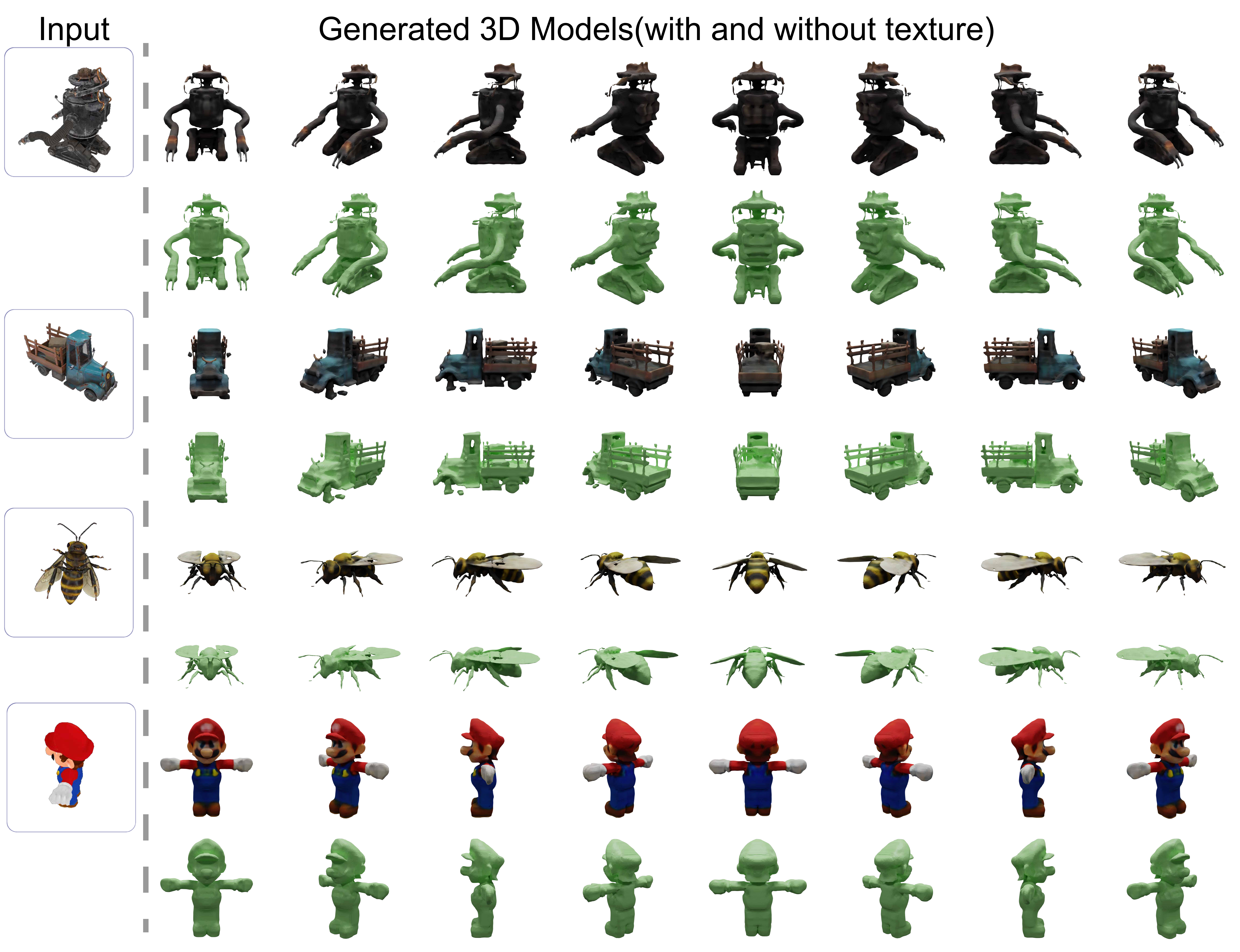}
    \caption{Given a single-view image, our method can generate high-quality 3D Models.}  
    \label{fig:teaser}
    \vspace{-1cm}
\end{figure}

%% file: aticle_architectures/abstract.tex
\begin{abstract}
  3D generation has witnessed significant advancements, yet efficiently producing high-quality 3D assets from a single image remains challenging. In this paper, we present a triplane autoencoder, which encodes 3D models into a compact triplane latent space to effectively compress both the 3D geometry and texture information. Within the autoencoder framework, we introduce a 3D-aware cross-attention mechanism, which utilizes low-resolution latent representations to query features from a high-resolution 3D feature volume, thereby enhancing the representation capacity of the latent space. Subsequently, we train a diffusion model on this refined latent space.
  In contrast to solely relying on image embedding for 3D generation, our proposed method advocates for the simultaneous utilization of both image embedding and shape embedding as conditions. Specifically, the shape embedding is estimated via a diffusion prior model conditioned on the image embedding. Through comprehensive experiments, we demonstrate that our method outperforms state-of-the-art algorithms, achieving superior performance while requiring less training data and time.
  Our approach enables the generation of high-quality 3D assets in merely 7 seconds on a single A100 GPU. More results and visualization can be found on our project page: \href{https://compress3d.github.io/}{https://compress3d.github.io/}.
  \keywords{3D Generation \and Diffusion Model}
\end{abstract}

%% file: aticle_architectures/introduction.tex
\section{Introduction}
3D assets are widely used and have huge demand in the fields of gaming, AR/VR, and films. However, 3D modeling is a time-consuming and labor-intensive job and requires a long period of learning and mastering a variety of tools. Although there are already some image generation algorithms that can assist designers in 3D modeling, directly generating high-quality 3D assets is still challenging.

Benefiting from the emergence of the large-scale image-text pairs dataset LAION, image generation algorithms have made great progress in both generation quality and diversity. DreamFusion\cite{poole2022dreamfusion} proposed score distillation sampling(SDS) for the first time, and used pre-trained 2D diffusion models to guide the generation of 3D models. Subsequent works replace the 3D scene representation with DMtet or Gaussian Splatting and improve the optimization process, which speeds up the generation process and improves the mesh quality. Learning-based 3D generation is also a promising direction, and our method also falls into this category. There have been some works\cite{jun2023shap, gupta20233dgen, mercier2024hexagen3d} training latent diffusion models on large-scale 3D datasets, achieving impressive results. However, none of these methods has a highly compressed latent space, which reduces the training speed and generation speed of latent diffusion. Moreover, current 3D generation methods use text or images as conditions to directly generate 3D models. However, these generated models usually do not conform to text or images, and the generated mesh geometry is low-quality.

\yty{To tackle the problems above, we propose a triplane autoencoder that takes colored point clouds as input to compress 3D model into a low-dimensional latent space on which a two-stage diffusion model is trained to generate 3D contents.}
\cite{mercier2024hexagen3d, gupta20233dgen} directly project 3D point-wise features to triplanes through mean pooling. \yty{As this process involves no learnable parameters, it inevitably leads to the loss of 3D information.} \cite{mercier2024hexagen3d, gupta20233dgen} use UNet to further refine the triplane, \yty{which however greatly increases computation due to the high-resolution of triplanes.}
We instead add learnable parameters in the process of \yty{projecting 3D point cloud to 2D triplanes, which mitigates the information loss while avoiding significant computational overhead}. Specifically, we first convert 3D point-wise features into 3D feature volume and then use 3D convolution neural networks in 3 directions to obtain high-resolution triplane features. We use a series of ResNet blocks and downsample layers to get a low-resolution triplane. To further enhance the representation ability of latents, Shap-E\cite{jun2023shap} uses multi-view images as additional input and injects multi-view information via cross-attention. \yty{However, multi-view images lack accuracy in representing 3D information and computing attention weights between image patch embeddings and latent tokens consumes significant time, resulting in inefficiency in encoder training.} In contrast, we leverage a 3D feature volume to augment the representation capability of triplane features. Specifically, we use triplane latent to query the 3D feature volume. This operation constitutes a local cross-attention mechanism that not only facilitates rapid computation but also significantly enhances the expressive capacity of triplane features.

\yty{Recovering 3D model from a single-view image is inherently an ill-posed problem. Instead of solely relying on image embedding for generating 3D, we propose leveraging both image embedding and shape embedding as conditions simultaneously for 3D content generation.} 
\yty{Shape embedding inherently contains more 3D information compared to image embedding. Therefore, incorporating shape embedding as an additional condition for 3D generation is expected to yield better results than conditioning solely on image embedding.}
To obtain shape embedding during generation, we train a diffusion prior model to generate shape embedding conditioned on the image embedding. Specifically, we first use a pre-trained shape-text-image alignment model OpenShape\cite{liu2024openshape} to extract the shape embedding of 3D model and the image embedding of its corresponding rendering image. We then train a diffusion prior model \yty{that can estimate shape embedding conditioned on the corresponding image embedding. Since these embeddings are aligned in the same space, it is easy to learn a model to convert image embedding into shape embedding.} 
Finally, we train a triplane latent diffusion model to generate triplane latent conditioned on the image embedding and the predicted shape embedding.

To summarize, our contributions are:
\begin{itemize}
    \item We design an autoencoder capable of efficiently compressing 3D models into a low-dimensional triplane latent space and accurately decoding them back to high-quality colored 3D models.
    \item We introduce a triplane latent diffusion model that can be conditioned on both image embeddings and shape embeddings estimated from image embeddings, thereby facilitating the generation of 3D models.
    \item We conduct extensive ablations studies to verify the effectiveness of different components of our method and demonstrate that our method achieves high-quality 3D generation from a single image.
\end{itemize} 



%% file: aticle_architectures/related_work.tex
\section{Related Work}
\subsection{Optimization-based Methods}
Different from image generation, the size of datasets for 3D generation is much smaller than that of 2D generation. The largest 3D dataset Objaverse-XL\cite{deitke2024objaverse} contains 10 million 3D objects, which is far smaller than LAION\cite{schuhmann2022laion} that is used to train text-to-image generation models. 
To alleviate the problem of lacking 3D data, DreamFusion\cite{poole2022dreamfusion} proposes score distillation sampling (SDS), which enables the use of a 2D pre-trained diffusion model as a prior for 3D optimization. However, the optimization process takes around 2 hours for one 3D asset. Make-it-3D\cite{tang2023make} incorporates constrain in the reference image and employs a two-stage optimization to achieve high-quality 3D generation. 
Magic3D\cite{lin2023magic3d} also adopts coarse to fine two-stage optimization, and it replaces the 3D scene representation from NeRF\cite{mildenhall2021nerf} to DMTet\cite{shen2021deep} in the refining stage, which allows it to efficiently render high-resolution images, greatly speeding up the optimization process and reducing the optimization time from 2 hours to 40 minutes. 
Recently, with the emergence of a new 3D scene representation Gaussian Splatting\cite{kerbl20233d}, there are also some works\cite{tang2023dreamgaussian, chen2023text, yi2023gaussiandreamer} that introduce this 3D representation into the field of optimization-based 3D generation. However, generating high-quality 3D assets using these optimization-based methods still takes several minutes.

\subsection{Learning-based Methods}
Limited by the scale of the 3D dataset, early learning-based 3D generation methods were limited to generating 3D geometry only. 
And there has been a large number of methods tried to explore generating point clouds\cite{wu2023fast, zeng2022lion, li2021sp}, mesh\cite{liu2023meshdiffusion, nash2020polygen, siddiqui2023meshgpt} and signed distance field(SDF)\cite{zhang20223dilg, shue20233d, zhang20233dshape2vecset, mittal2022autosdf, shim2023diffusion, li2023diffusion, cheng2023sdfusion}. Due to its sparse nature, point clouds are difficult to reconstruct fine 3D geometry. Computing the signed distance field requires preprocessing of the 3D mesh, and the geometry quality of the processed mesh will decrease. With the emergence of new 3D scene representations (NeRF\cite{mildenhall2021nerf}, DMTet\cite{shen2021deep}, Gaussian Splatting\cite{kerbl20233d}, FlexiCubes\cite{shen2023flexible}) and large-scale 3D datasets, it is possible to replicate the successes of image generation in the field of 3D generation.
Point-E\cite{nichol2022point} train a diffusion transformer with CLIP\cite{radford2021learning} image embedding as a condition on a large-scale 3D dataset to generate coarse colored point cloud, and then use a point cloud upsampler to upsamle coarse colored point cloud. Compared to optimization-based 3D generation methods, it is one to two orders of magnitude faster to sample from. However, since the generated point cloud contains only 4K points, it is difficult to reconstruct high-quality 3D mesh.
To generate high-quality 3D mesh, Shpa-E\cite{jun2023shap} uses a transformer encoder to encode colored point cloud and multi-view images into parameters of an implicit function, through which mesh and neural radiance fields can be generated. Shpa-E then trains a conditional latent diffusion transformer to generate the parameters of the implicit function. Shap-E demonstrates the potential of latent representation in the field of 3D generation. Subsequent works\cite{gupta20233dgen, mercier2024hexagen3d} also train the diffusion model on the latent space, but use DMTet\cite{shen2021deep} as the 3D scene representation, which improves the training speed and geometry quality. However, how to compress 3D model into a low-dimensional latent space is still an open problem.

\subsection{Reconstruction-based Methods}
There are also some methods that use 3D reconstruction techniques to generate 3D assets.
Zero-1-to-3\cite{liu2023zero} proposes that for a single-view image of a given object, images of other specific views of the object are generated through fine-tuning a 2D diffusion model, and then reconstruct 3D assets through the generated multi-view images. One-2-3-45\cite{liu2024one} further improves view consistency and reconstruction efficiency. LRM\cite{hong2023lrm} and Instant3d\cite{li2023instant} use a transformer to encode images into a triplane and use NeRF to reconstruct the 3D assets.
Some recent work has introduced the gaussian splatting technique into the field of reconstruction-based 3D generation to achieve more efficient and high-quality reconstruction. \cite{zou2023triplane} uses a hybrid triplane-gaussian intermediate representation for single-view reconstruction that efficiently generates a 3D model from a single image via feed-forward inference. More recently, LGM\cite{tang2024lgm} proposes to encode multi-view images into multi-view gaussian features for high-quality 3D model generation.

%% file: aticle_architectures/method.tex
\input{figures/overview}

\section{Method}
Our approach uses latent diffusion models to generate 3D assets from a single image. Instead of generating on the latent space 
of 3D models directly, we first generate shape embedding conditioned on the image embedding, then we generate triplane latent conditioned on both image embedding and previously generated shape embedding. The overview of our method is shown in Fig.~\ref{fig:overview}.

Specifically, our method consists of three stages. In the first stage, we train a triplane variational autoencoder which takes as input the colored point clouds. The triplane encoder encodes 3D geometry and texture on a compressed triplane latent space. Subsequently, a triplane decoder reconstructs colored 3D model from the triplane latent space.
In the second stage, we train a diffusion prior model to generate shape embedding conditioned on the image embedding. To obtain shape and image embedding pairs, we use OpenShape\cite{liu2024openshape} to extract the shape embedding of 3D model and the image embedding of its rendered image. 
In the third stage, we train a triplane diffusion model to generate triplane latent conditioned on the image embedding and shape embedding.

At the inference time, our method takes a single image as input. We utilize the CLIP model \cite{radford2021learning} to extract the image embedding. Next, we employ the diffusion prior model to generate shape embedding conditioned on the image embedding. Then, using a triplane diffusion network, we condition on the previously generated shape embedding and image embeddings to generate triplane latent, which is finally decoded into a colored 3D model.

\subsection{Triplane AutoEncoder}

\subsubsection{Encoder}
\label{sec:encoder}
The triplane encoder is shown in Fig.~\ref{fig:encoder}. The encoder takes colored point clouds as input and outputs a distribution on the triplane latent space. 
We represent the colored point cloud as $P \in \mathbb{R}^{N\times6}$, where $N$ is the number of points. The first three channels represent point 3D position $(x, y, z)$ and the last three channels represent its corresponding $(R, G, B)$ colors. We use PointNet\cite{qi2017pointnet} with position embedding and local max pooling as our point cloud encoder to extract 3D point-wise features. Then we project 3D point-wise features onto triplanes to achieve feature compression. 

\input{figures/encoder}

Previous methods\cite{mercier2024hexagen3d, gupta20233dgen} directly project 3D point-wise features to triplanes through mean pooling, which inevitably leads to the loss of 3D information due to no learnable parameters in this process. Other works, such as 3DGen \cite{gupta20233dgen}, employ a UNet to further refine the triplane features and mitigate the loss of 3D information. However, incorporating an additional UNet does not compress the triplane and may increase computational demands.
We instead add learnable parameters in this process. Specifically, given point-wise features $F =\{f_i\in \mathbb{R}^{c} \}_N$, the feature volume $V =\{v_j \in \mathbb{R}^{c} \}_{r\times r\times r} \in \mathbb{R}^{r \times r \times r \times c}$ is calculated as
\begin{equation}
    v_j = \sum_{i\in \mathcal{N}(j)}{w_i \cdot f_i}
    \label{eq:feat_volume}
\end{equation}
where $r$ is the resolution of feature volume and $c$ is the number of channels. $\mathcal{N}(j)$ is a set which contains the neighbor points indices of the $j$th feature volume grid, and $w_i=(1-|p_j^x-p_i^x|)(1-|p_j^y-p_i^y|)(1-|p_j^z-p_i^z|)$ is a weight that is inversely proportional to the distance between $p_i$ and $p_j$. The 2D illustration of the conversion is shown in Fig.~\ref{fig:encoder}. 
As the point cloud density is usually uneven, we need to normalize $v_j$ to cancel out the impact of the point cloud density. We obtain the normalized feature volume $V^{n} = \{v^{n}_j \in \mathbb{R}^{c} \}_{r\times r\times r} \in \mathbb{R}^{r \times r \times r \times c}$ by,

\begin{equation}
    v_j^{n} = \frac{v_j}{\sum_{i\in N(j)}{w_i}}
    \label{eq:feat_volume_norm}
\end{equation}

After obtaining normalized feature volume $V^{n}$, We employ 3D convolution in three directions to convolve the normalized feature volume and obtain high-resolution triplane features $T_{xy}, T_{yz}, T_{zx}\in \mathbb{R}^{r\times r\times c}$, respectively. 
\begin{align}
    T_{xy} &= \texttt{3DConv}(V^{n}, k=(1, 1, r), s=(1, 1, r)) \\
    T_{yz} &= \texttt{3DConv}(V^{n}, k=(r, 1, 1), s=(r, 1, 1)) \\
    T_{zx} &= \texttt{3DConv}(V^{n}, k=(1, r, 1), s=(1, r, 1))
\end{align}
where $k$ is the kernel size and $s$ is the stride.
Then the triplane features are passed through a series of ResBlocks and down-sample layers to obtain low-resolution triplane latents $T_{xy}^{l}$, $T_{yz}^{l}$, $T_{zx}^{l} \in  \mathbb{R}^{r^{\prime}\times r^{\prime} \times c^{\prime}}$. 

To enhance the representation ability of triplane latents, we propose a 3D-aware cross-attention mechanism, which takes triplane features as queries to query features from 3D feature volume. The 3D-aware cross-attention computation process is shown in Fig.~\ref{fig:cross_attention}. 
We first use a 3D convolutional layer to down sample $V^{n}$ to obtain a low-resolution feature volume $V^{n}_{d} \in \mathbb{R}^{r^{\prime\prime}\times r^{\prime\prime}\times r^{\prime\prime}\times c^{\prime\prime}}$. 
\begin{equation}
    V^{n}_{d} = \texttt{3DConv}(V^{n}, k=(o, o, o), s=(o, o, o))
\end{equation}
where $o$ is the down-sample factor. Then, leveraging low-resolution triplane latents $T_{xy}^{l}$, $T_{yz}^{l}$, and $T_{zx}^{l}$, we employ 3D-aware cross-attention on the feature volume $V^{n}_{d}$ to extract a residual feature. This residual feature is then added to the original triplane latent to compose the enhanced triplane latent.
\begin{equation}
    ({T_{xy}^{{e}}}, {T_{yz}^{{e}}}, {T_{zx}^{{e}}}) = (A_{xy},A_{yz}, A_{zx}) + (T_{xy}^{l}, T_{yz}^{l}, T_{zx}^{l})
\end{equation}
where ${T_{xy}^{{e}}}$, ${T_{yz}^{{e}}}$, ${{T_{zx}^{{e}}}}$ are enhanced triplane latents. $A_{xy}$, $A_{yz}$, $A_{zx}$ are the residual feature obtained by 3D-aware cross-attention. We empirically found that querying on low-resolution feature volume does not hurt the performance while saving lots of computation as shown in Table \ref{table:ablation_for_feature_volume_size_in_cross_attention}. To compute the residual features, we need first calculate the triplane queries $Q_{xy}$, $Q_{yz}$, $Q_{zx} \in  \mathbb{R}^{r^{\prime}\times r^{\prime} \times d}$ and feature volume keys $K\in  \mathbb{R}^{r^{\prime\prime}\times r^{\prime\prime}\times r^{\prime\prime} \times d}$ and values $V\in  \mathbb{R}^{r^{\prime\prime}\times r^{\prime\prime}\times r^{\prime\prime}\times c^{\prime}}$ by, 
\begin{equation}
\begin{split}
    (Q_{xy}, Q_{yz}, Q_{zx}) & = \texttt{TriConv}((T_{xy}^{l}, T_{yz}^{l}, T_{zx}^{l}), k=(1,1), s=(1,1)) \\
    K &= \texttt{3DConv}(V^{n}_{d}, k=(1,1,1), s=(1,1,1)) \\
    V &= \texttt{3DConv}(V^{n}_{d}, k=(1,1,1), s=(1,1,1)) \\
\end{split}
\end{equation}
where $\texttt{TriConv}$ is the 3D-aware convolution proposed in\cite{wang2023rodin}. For simplicity, we take $A_{xy}$ as an example to illustrate 3D-aware cross-attention process. $A_{yz}, A_{zx}$ can be calculated in a similar way. 
We define $Q_{xy}=\{q_{ij}\in \mathbb{R}^{1 \times d}\}_{r^{\prime}\times r^{\prime}}$ where $q_{ij}$ is one point feature at position $(i,j)$. We then extract its corresponding key and value by,
\begin{align}
    k_{ij} &= K(mi:mi+m-1,mj:mj+m-1,:,:) \in \mathbb{R}^{m\times m \times r^{\prime\prime} \times d} \\
    v_{ij} &= V(mi:mi+m-1,mj:mj+m-1,:,:) \in \mathbb{R}^{m\times m \times r^{\prime\prime} \times c^{\prime}}
\end{align}
where $m = \texttt{round}(\frac{r^{\prime\prime}}{r^{\prime}})$ is the scale ratio between volume size and triplane size. We then reshape $k_{ij}$ and $v_{ij}$ to $\mathbb{R}^{m^2r^{\prime\prime} \times d}$ and $\mathbb{R}^{m^2r^{\prime\prime} \times c^{\prime}}$ repectively for ease of attention computation.
The cross-attention feature $A_{xy}=\{a_{ij} \in \mathbb{R}^{1 \times c^{\prime}}\}_{r^{\prime}\times r^{\prime}}$ can be calculated by,
\begin{equation}
    a_{ij} = \texttt{sotfmax}(\frac{q_{ij}k_{ij}^{T}}{\sqrt{d}})v_{ij}
\end{equation}

 \input{figures/3d_aware_cross_attention}




\input{figures/decoder}

\subsubsection{Decoder}
As shown in Fig.~\ref{fig:decoder}, the decoder consists of a series of ResBlocks and up-sample layers. The decoder is responsible for decoding the low-resolution triplane latent into a high-resolution triplane feature. The high-resolution triplane feature contains the geometry and texture information of the 3D model. 

To recover geometry information from triplane features, we adopt FlexiCubes\cite{shen2023flexible} representation, an isosurface representation capable of generating high-quality mesh with low-resolution cube grids. For each cube in FlexiCubes, we predict the weight, signed distance function (SDF), and vertex deformation at each cube vertex. Specifically, we concatenate the triplane features of the eight vertices of each cube and predict the cube weight using an MLP layer. Similarly, we concatenate the triplane features of each vertex to predict the SDF and deformation using another 2 MLP layers. With the cube weights, SDF, and vertex deformations determined, the mesh can be extracted using the dual marching cubes method\cite{schaefer2004dual}. 
To recover texture information from the triplane features, we take the triplane features of the mesh surface points and predict the color of each surface point through an MLP layer.

\subsubsection{Renderer}
We train the encoder and decoder using a differentiable renderer\cite{Laine2020diffrast}. Compared with previous methods\cite{mittal2022autosdf, zhang20223dilg, zhang20233dshape2vecset}, we do not need to pre-compute the signed distance field of each 3D mesh, which demands a huge computation and storage space. Moreover, our method based on differentiable rendering also avoids information loss during data pre-processing. For the mesh output by the decoder, we first render the 3D model at a certain view and then compare it with the rendering images of the ground truth model from the same perspective. The rendering images contains RGB image $I_{rgb}$, silhouette image $I_{mask}$ and depth image $I_{depth}$. 
Finally, we calculate the loss in the image domain and train the encoder and decoder jointly through rendering loss $L_{R}$. The rendering loss is as follows:

\begin{equation}
    L_{R} = \lambda_{1}L_{rgb} + \lambda_{2}L_{mask} + \lambda_{3}L_{depth} - \lambda_{kl}D_{KL}(N(\mu, \sigma) | N(0,1))
\end{equation}
where $L_{rgb}=||I_{rgb}-I_{rgb}^{gt}||^2$, $L_{mask}=||I_{mask}-I_{mask}^{gt}||^2$, $L_{depth}=||I_{depth}-I_{depth}^{gt}||^2$, $N(\mu, \sigma)$ is the distribution of the low resolution triplane latent.
Moreover, we add KL penalty to ensure that the distribution of the triplane latent $N(\mu, \sigma)$ is close to the standard Gaussian distribution $N(0,1)$. 

\subsection{Diffusion Prior Model}
Generating a 3D model directly from an image is a difficult task because the image embedding of a single view image only contains 2D geometry and texture information of the 3D model. Compared to image embedding, shape embedding contains richer 3D geometry and texture information. Generating 3D model with shape embedding as a condition is easier and more accurate than using image embedding as a condition.
To train this diffusion prior model, we first use the OpenShape\cite{liu2024openshape} model pre-trained on large-scale 3D dataset, a shape-text-image alignment model, to extract the shape embedding $e_s\in \mathbb{R}^{1280}$ of the 3D model and the image embedding $e_i\in \mathbb{R}^{1280}$ of the single-view rendering image.
Then we design an MLP with skip connections between layers at different depths of the network as the diffusion backbone to generate shape embedding. 
This diffusion backbone consists of multiple MLP ResBlocks. In each block, image embedding is injected into the MLP block through concatenation, and the timestep embedding is injected through addition.
Instead of using $\epsilon$-prediction formulation as used in \cite{ho2020denoising}, we train our prior diffusion model to predict the denoised $e_s$ directly with 1000 denoising steps, and use a L1 loss on the prediction:
\begin{equation}
    L_{prior} = \mathbb{E}_{t\sim[1,T], e_s^{(t)}\sim q_t}[|| f^{p}_{\theta}( e_s^{(t)}, t, e_i)-e_s ||]
\end{equation}
where $f^{p}_{\theta}$ is the learned prior model.

\subsection{Triplane Diffusion Model}
\label{sec: triplane_latent_diffusion}
After we obtain the prior model, we then train a triplane diffusion model, which uses the shape embedding estimated by the prior model and image embedding as conditions, to generate 3D models. The diffusion backbone is a UNet, which contains multiple ResBlocks and down/up sample layers. The input and output of each ResBlock are triplanes, and we use 3D-aware convolution\cite{wang2023rodin} in each ResBlock. Shape embedding $e_s$ and image embedding $e_p$ are injected into ResBlocks through cross attention.
We train the triplane diffusion model to predict the noise $\epsilon$ added to the triplane latent with 1000 denoising steps, and use an L1 loss on the prediction,
\begin{equation}
    L_{tri} = \mathbb{E}_{t\sim[1,T], \epsilon\sim N(0,1)}[|| f_{\theta}(z^{t}, t, e_s, e_p)-\epsilon ||]
\end{equation}
where $f_{\theta}$ is the learned triplane diffusion model. 
To improve the diversity and quality of generated samples, we introduce classifier free guidance\cite{ho2022classifier} by randomly dropout conditions during training. Specifically, we randomly set only $e_p=\varnothing_{p}$ for $5\%$, only $e_s=\varnothing_{s}$ for $5\%$, both $e_p=\varnothing_{p}$ and $e_s=\varnothing_{s}$ for $5\%$. During the inference stage, the score estimate is defined by,

\begin{equation}
\begin{split}
    \widetilde{f_{\theta}}(z^{t}, t, e_s, e_p) =& f_{\theta}(z^{t}, t, \varnothing_{s}, \varnothing_{p}) \\
       &+ s_p \cdot (f_{\theta}(z^{t}, t, \varnothing_{s}, e_{p}) - f_{\theta}(z^{t}, t, \varnothing_{s}, \varnothing_{p})) \\
       &+ s_s \cdot (f_{\theta}(z^{t}, t, e_{s}, e_{p}) - f_{\theta}(z^{t}, t, \varnothing_{s}, e_{p}))
\end{split}
\end{equation}

%% file: figures/overview.tex
\begin{figure}[t]
    \centering
    \includegraphics[width=0.90\textwidth]{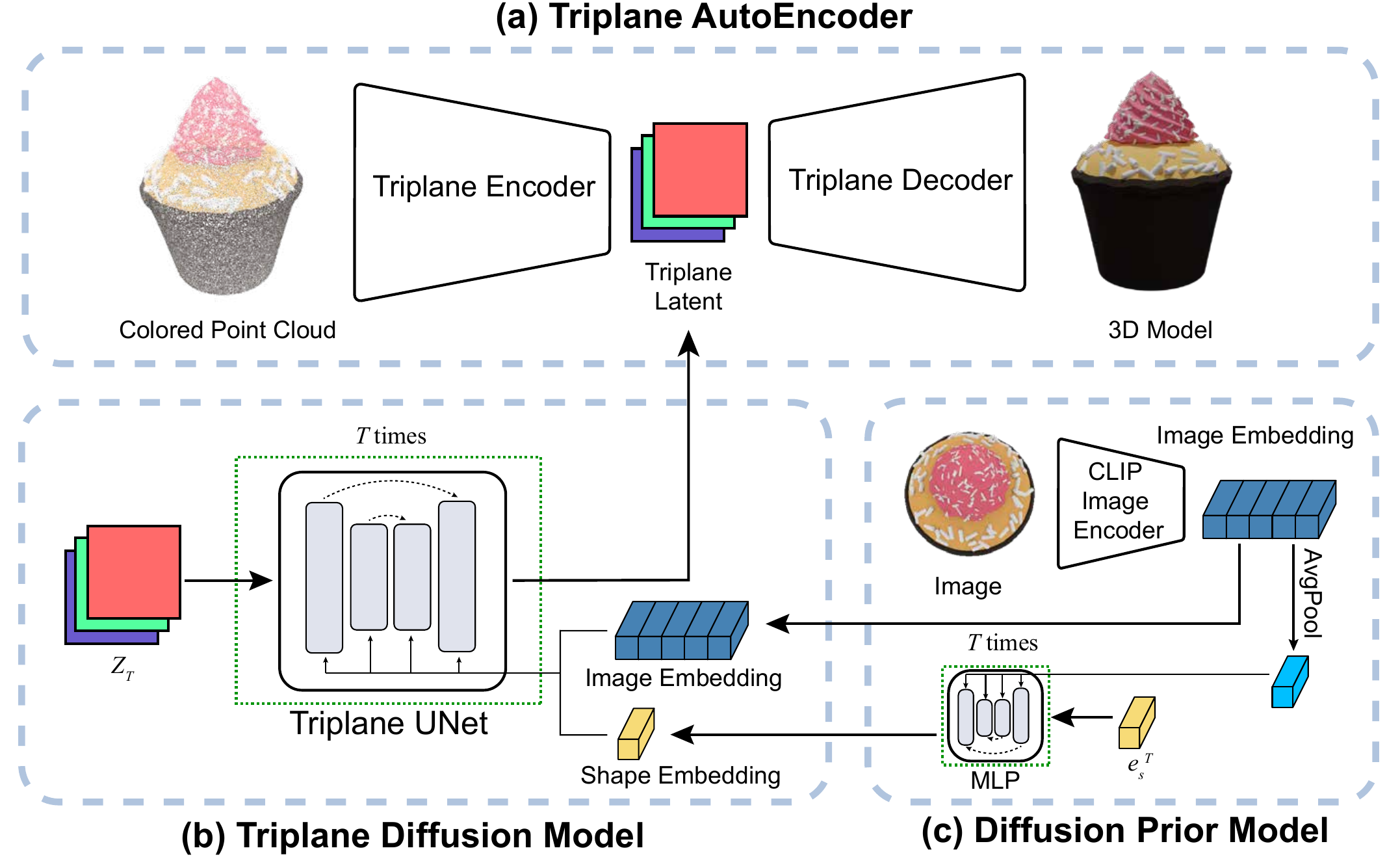}
    \caption{Method overview. Compress3D mainly contains 3 components. (a) Triplane AutoEncoder:  Triplane Encoder encodes color point cloud on a low-resolution triplane latent space. Then we use a Triplane Decoder to decode 3D model from a triplane latent. (b) Triplane Diffusion Model: we use shape embedding and image embedding as conditions to generate triplane latent. (c) Diffusion Prior Model: generate shape embedding conditioned on the image embedding. }  
    \label{fig:overview}
    \vspace{-0.5cm}
\end{figure}

%% file: figures/encoder.tex
\begin{figure}[t]
    \centering
    \includegraphics[width=\textwidth]{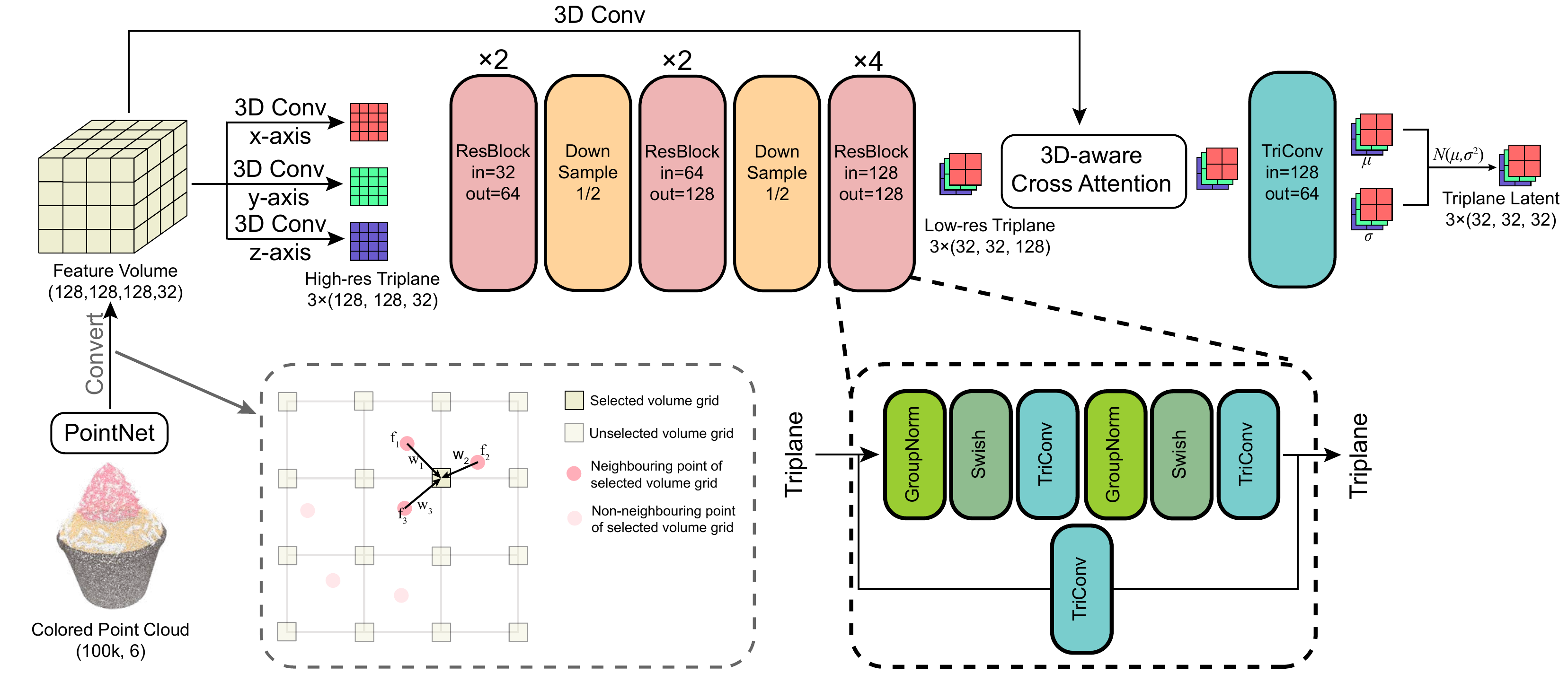}
    \caption{Triplane Encoder. TriConv is the 3D-aware convolution proposed in \cite{wang2023rodin}.}  
    \label{fig:encoder}
    \vspace{-0.5cm}
\end{figure}

%% file: figures/3d_aware_cross_attention.tex
\begin{figure}[t]
    \vspace{0cm}
    \centering
    \includegraphics[width=0.5\textwidth]{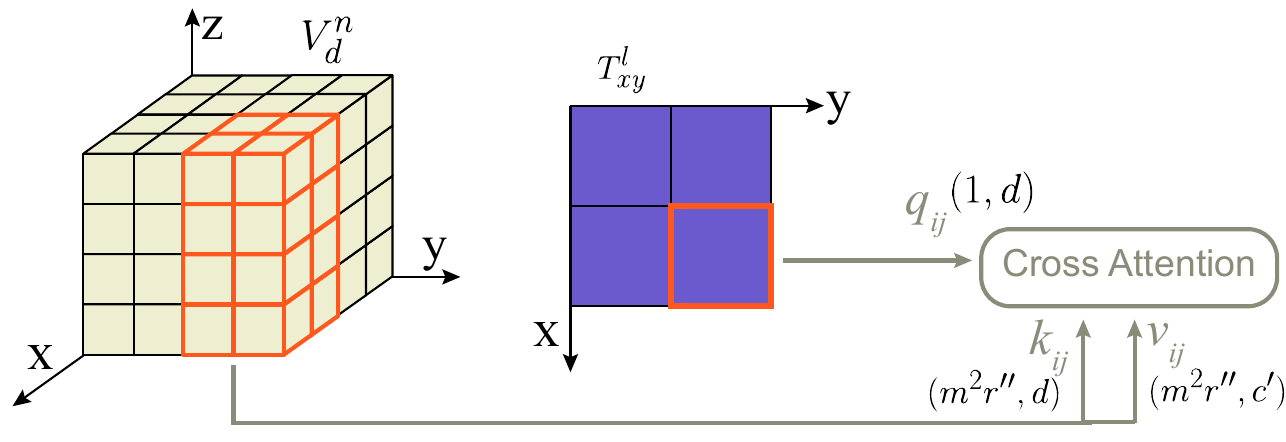}
    \caption{3D-aware cross attention. We use each point feature on the triplane to query the corresponding cube region (red) of feature volume. In addition, we add a position embedding to the volume feature.}  
    \label{fig:cross_attention}
    \vspace{-0.5cm}
\end{figure}

%% file: figures/decoder.tex
\begin{figure}[htb]
    \centering
    \includegraphics[width=0.8\textwidth]{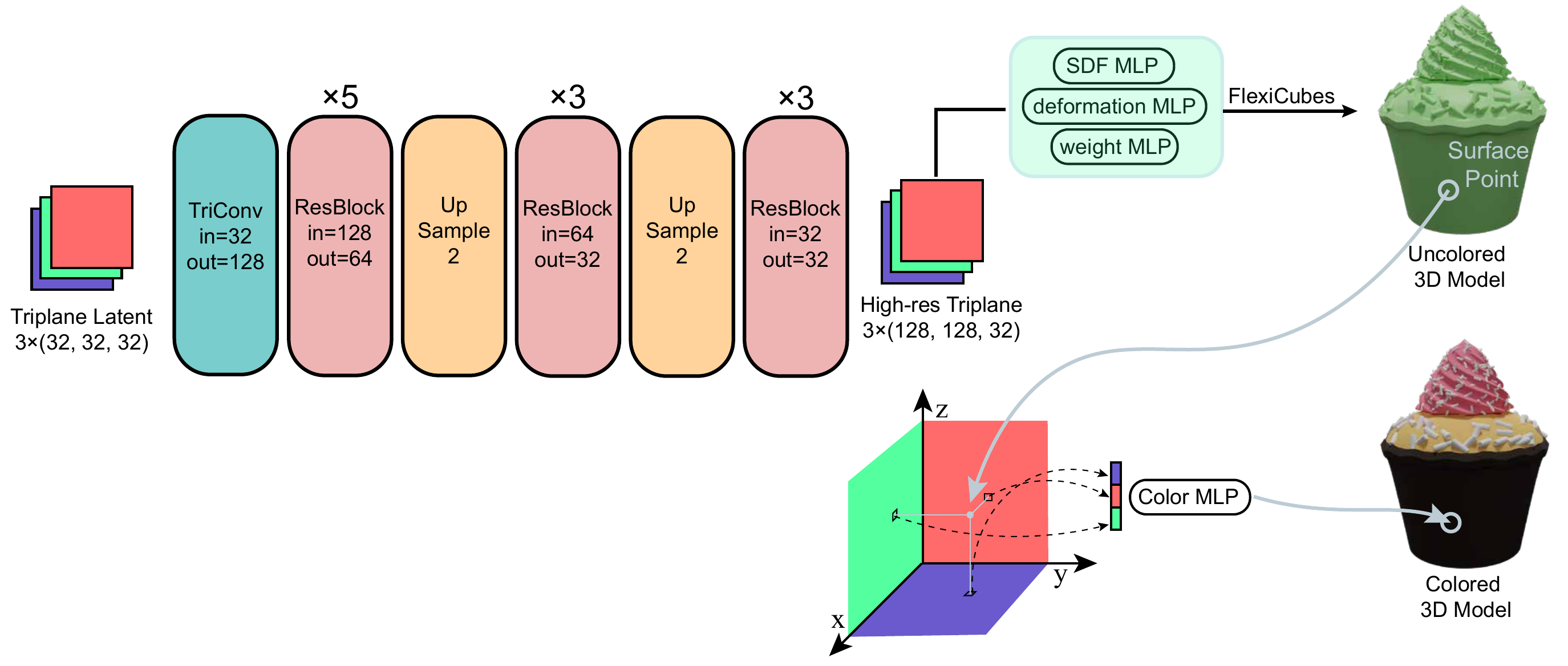}
    \caption{Triplane Decoder. }  
    \label{fig:decoder}
    \vspace{-0.5cm}
\end{figure}

%% file: aticle_architectures/experiments.tex
\section{Experiments}
\subsection{Dataset Curation}
We train our model on a filtered Objaverse dataset\cite{deitke2023objaverse}. As there are many low-quality 3D models in the origin Objaverse dataset. To obtain high-quality 3D data for training, we manually annotated approximately 2500 3D models, categorizing them as either good or bad. A 'good' 3D model exhibits realistic textures and intricate geometric structures, whereas a 'bad' 3D model is characterized by single-color textures or simple shapes. We randomly select five random views and use the pre-trained CLIP model to extract their image embeddings. Then we concatenate these image embeddings and feed them into a shallow MLP network for classification. Despite the limited annotation data, we find that the trained MLP classification network can correctly classify 3D models in most cases. We use this MLP classification network to filter the entire Objaverse dataset and obtain 100k high-quality 3D models. We randomly select 95\% 3D models for training and 5\% for testing.

\subsection{Training Details}
\textbf{Triplane AutoEncoder} 
For the encoder, the number of input points $N$ is 100k, the resolution $r$ of the $V_{norm}$ is 128, the resolution $r^{\prime\prime}$ of the $V_{d}^{n}$ used in 3D-aware cross attention is 32. The resolution $r^{\prime}$ of the triplane latent is 32, and its channel number is 32. For the decoder, the decoded triplane has a resolution of 128, and its channel number is 32, we set the grid size of FlexiCubes as 90. For the Renderer, we render $512\times512$ RGB, mask and depth images from 40 random views to supervise the training process, and we set $\lambda_1=10$, $\lambda_2=10$, $\lambda_3=0.1$, $\lambda_{kl}=1e^{-6}$ for the rendering loss. The triplane autoencoder has 32M parameters in total, and it is trained with the AdamW optimizer. The learning rate gradually decreases from $3\times10^{-5}$ to $3\times10^{-6}$. We train it on 8 A100 GPUs for 6 days.

\noindent\textbf{Diffusion Prior Model}
To stabilize the training process of the prior diffusion network, we scale the shape embedding $e_s$ by 0.25, and image embedding $e_i$ by 0.85, making their variance approximate to 1. The Diffusion Prior Model has 25.8M parameters, and we train it on 2 A100 GPUs for 18 hours. The learning rate gradually decreases from $1\times10^{-5}$ to $1\times10^{-6}$. 

\noindent\textbf{Triplane Diffusion Model}
The triplane diffusion model has 864M parameters, We train the model on 8 A100 GPUs for 4 days. The learning rate gradually decreases from $3\times10^{-5}$ to $3\times10^{-6}$.

\input{tables/fid_clip_table}

\subsection{Comparison with Other Methods}
We compare our method with Shap-E\cite{jun2023shap} and OpenLRM\cite{openlrm}. To generate 3D model efficiently, We use DDIM\cite{song2020denoising} sampler with 50 steps. The guidance scale for shape embedding and image embedding are 1.0 and 5.0 respectively. 

\noindent\textbf{Quantitative Comparison} 
We use FID and CLIP similarity as evaluation metrics for generation quality. For the computation of FID, we randomly select 200 images in our test set that have not been seen during training, and generate 200 3D models using our method. Then we render each generated 3D model and its corresponding ground truth 3D model from 40 random views. We compute FID of the generated images set and ground truth image set. For the CLIP similarity, we calculate the cosine similarity of the CLIP image embedding of the generated 3D model and GT 3D model at the same viewpoint. We calculate FID and CLIP similarity five times and take the average. 
The quantitative comparison is reported in Table~\ref{table:fid_clip}. Our method achieves lower FID and higher CLIP similarity than Shap-E and OpenLRM, while using less training data and time. 

\input{figures/compare_with_other_methods_figure}

\noindent\textbf{Qualitative Comparison}
The qualitative comparison is shown in Fig.~\ref{fig:compare_with_other_methods}. Compared with other methods, Compress3D can generate 3D models with good texture and fine geometric details. Benefiting from the two-stage generation, our method can generate high-quality results under various viewing angles, while OpenLRM and Shpa-E are more sensitive to viewing angles. For example, OpenLRM and Shpa-E usually fail to generate 3D models with fine geometric details given top and bottom views as input. In addition, the up-axis of the 3D model generated by OpenLRM often does not coincide with the z-axis, which needs to be manually rotated to align with the z-axis This is time-consuming and laborious. In comparison, our method could generate 3D models whose up-axis coincides with the z-axis, which makes it easier to use.

\subsection{Ablation Studies}
To evaluate the design of our method, we conduct a series of ablation studies on several key designs.

\noindent\textbf{3D-aware cross-attention.}
As described in Section~\ref{sec:encoder}, to enhance the representation ability of the triplane latent, we use triplane to query a feature volume via 3D-aware cross-attention. Table~\ref{table:ablation_for_using_cross_attention} shows that 3D-aware cross-attention improves the geometric and texture reconstruction quality greatly. Although the training time for each step increases slightly, from 0.789s to 0.824s, this is acceptable. 
As shown in Table~\ref{table:ablation_for_feature_volume_size_in_cross_attention}, we find using a down-sampled feature volume in 3D-aware cross-attention improves reconstruction quality slightly and greatly decreases the training time.  

\input{tables/ablation_for_using_cross_attention}

\input{tables/ablation_for_feature_volume_size_in_cross_attention}

\input{figures/compare_w_wo_shape_embedding}

\noindent\textbf{Diffusion Prior Model.}
To validate the importance of diffusion prior model, we train a triplane diffusion model conditioned only on the image embedding and compare it with our method. As shown in Table~\ref{table:ablation_for_using_shape_embedding}, using prior model further improves the quality of generated 3D model. As shown in Figure~\ref{fig:compare_w_wo_shape_embedding}, our method can still produce correct shapes under some unusual viewing angles, while the one without prior model fails.

\input{tables/ablation_for_using_shape_embedding}

\noindent\textbf{Guidance scales.}
To increase the quality of generated 3D models, we adopt classifier-free guidance during inference. There are multiple combinations of guidance scales for shape embedding and image embedding. Overall, we find that an appropriate guidance scale for $s_p$ or $s_s$ can improve the generation quality. As shown in Table \ref{table:ablation_for_guidance_scale}, when $s_p=5.0$, $s_s=1.0$, the model achieves the best FID. Although its CLIP similarity is slightly lower than the best one, they are very close. 

\input{tables/ablation_for_guidance_scale}

%% file: tables/fid_clip_table.tex
\setlength{\tabcolsep}{4pt}
\begin{table}[t]
\begin{center}
\caption{Quantitative Comparison with other methods.}
\label{table:fid_clip}
\begin{tabular}{l c c c}
\hline\noalign{\smallskip}
Metric & Shap-E\cite{jun2023shap} & OpenLRM\cite{openlrm} & Ours\\
\noalign{\smallskip}
\hline
\noalign{\smallskip}
FID($\downarrow$)            & 146.14	& 94.47 & \bf{53.21}\\
CLIP Similarity($\uparrow$)  & 0.731	& 0.756 & \bf{0.776}\\
Latent space dimension($\downarrow$) & 1.05M & 0.98M & \bf{0.10M}\\
Seconds per shape($\downarrow$) & 11 & \bf{5} & 7\\ 
Training dataset size & $\geq$1M & 0.951M & \bf{0.095M}\\ 
Training time (A100 GPU hours) & - & 9200 & \bf{1900}\\ 
\hline
\end{tabular}
\end{center}
\vspace{-0.7cm}
\end{table}
\setlength{\tabcolsep}{1.4pt}

%% file: figures/compare_with_other_methods_figure.tex
\begin{figure}[p]
    \vspace{-0.3cm}
    \centering
    \includegraphics[width=0.8\textwidth]{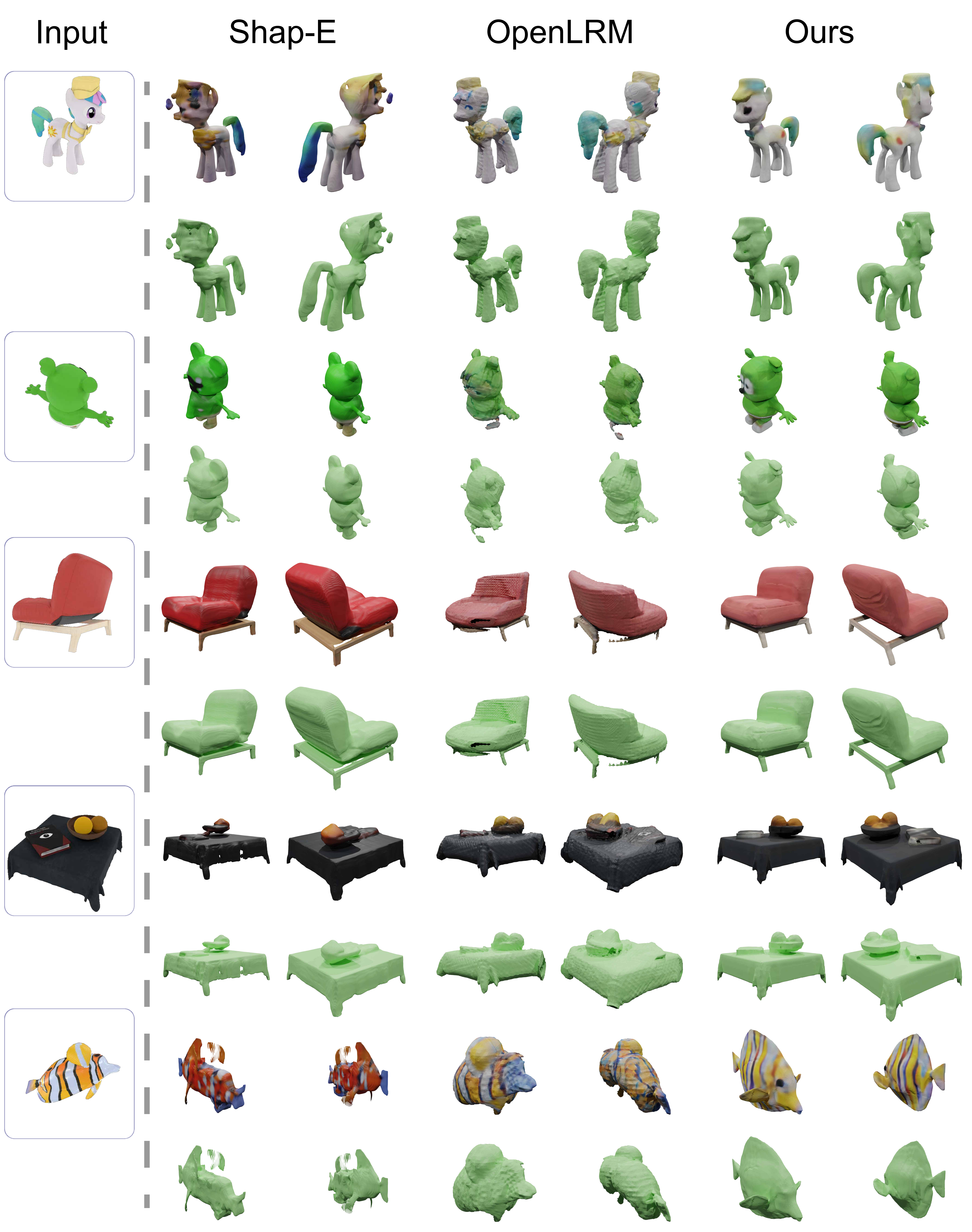}
    \caption{Qualitative comparison with other methods.}  
    \label{fig:compare_with_other_methods}
    \vspace{-0.3cm}
\end{figure}

%% file: tables/ablation_for_using_cross_attention.tex
\setlength{\tabcolsep}{4pt}
\begin{table}[h]
\vspace{0cm}
\begin{center}
\caption{Ablation study on 3D-aware cross attention.}
\label{table:ablation_for_using_cross_attention}
\begin{tabular}{l c c c c}
\hline\noalign{\smallskip}
Method & $L_{rgb}\times10^{3}$($\downarrow$) & $L_{mask}\times10^{3}$($\downarrow$) & $L_{depth}\times10^{2}$($\downarrow$) & seconds per step($\downarrow$)\\
\noalign{\smallskip}
\hline
\noalign{\smallskip}
w/o attention  & 3.798	    & 6.953    & 2.637   &\bf{0.789}\\
w attention     & \bf{2.485}	& \bf{5.059}& \bf{2.095} & 0.824\\
\hline
\end{tabular}
\end{center}
\vspace{0cm}
\end{table}
\setlength{\tabcolsep}{1.4pt}

%% file: tables/ablation_for_feature_volume_size_in_cross_attention.tex
\setlength{\tabcolsep}{4pt}
\begin{table}[htbp]
\vspace{0cm}
\begin{center}
\caption{Ablation study on volume resolution $r^{\prime\prime}$ used in 3D-aware cross attention.}
\label{table:ablation_for_feature_volume_size_in_cross_attention}
\begin{tabular}{l c c c c}
\hline\noalign{\smallskip}
Resolution & $L_{rgb}\times10^{3}$($\downarrow$) & $L_{mask}\times10^{3}$($\downarrow$) & $L_{depth}\times10^{2}$($\downarrow$) & seconds per step($\downarrow$)\\
\noalign{\smallskip}
\hline
\noalign{\smallskip}
128  & 2.551	    & 5.234      & 2.187        & 2.295\\
64   & 2.497	    & 5.134      & 2.124        & 0.961\\
32(ours)   & \bf{2.485}	& \bf{5.059} & \bf{2.095}  & \bf{0.824}\\
\hline
\end{tabular}
\end{center}
\vspace{-00cm}
\end{table}
\setlength{\tabcolsep}{1.4pt}

%% file: figures/compare_w_wo_shape_embedding.tex
\begin{figure}[thbp]
    \centering
    \includegraphics[width=0.8\textwidth]{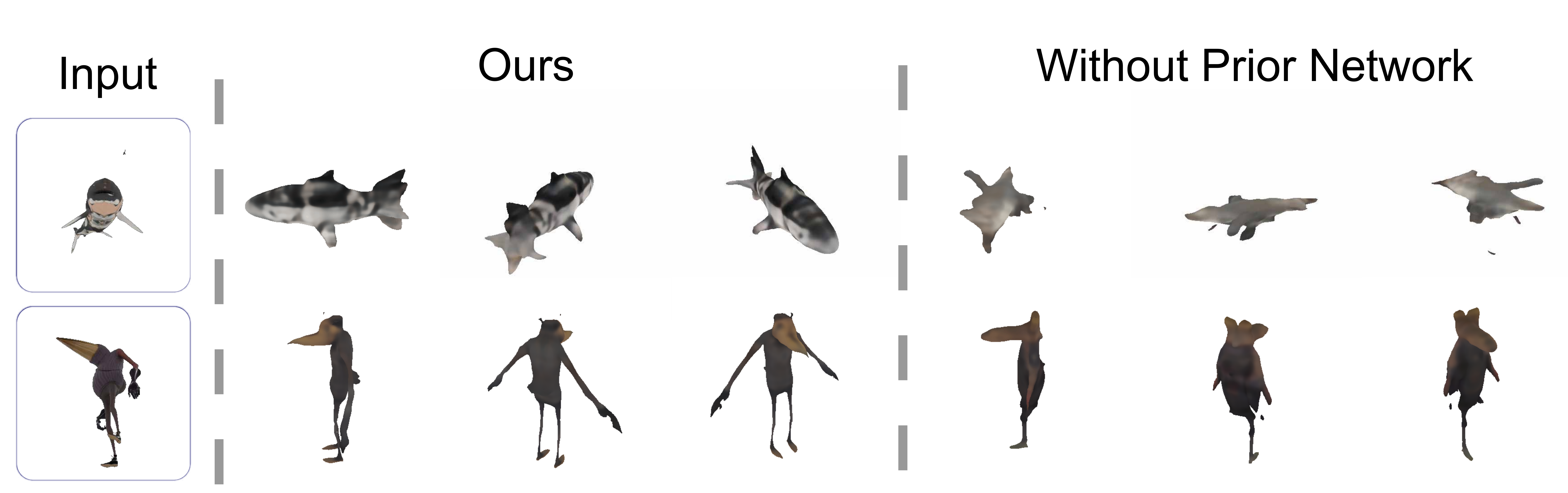}
    \caption{Ablation Study: Compare our method with the version that do not use prior diffusion network.}  
    \label{fig:compare_w_wo_shape_embedding}
    \vspace{-0.5cm}
\end{figure}

%% file: tables/ablation_for_using_shape_embedding.tex
\setlength{\tabcolsep}{4pt}
\begin{table}[h]
\vspace{0cm}
\begin{center}
\caption{Ablation study on using diffusion prior model.}
\label{table:ablation_for_using_shape_embedding}
\begin{tabular}{l c c}
\hline\noalign{\smallskip}
Method & FID($\downarrow$) & CLIP Similarity($\uparrow$)\\
\noalign{\smallskip}
\hline
\noalign{\smallskip}
w/o prior & 66.46	    & 0.745\\
w prior & \bf{53.21}	& \bf{0.776}\\
\hline
\end{tabular}
\end{center}
\vspace{0cm}
\end{table}
\setlength{\tabcolsep}{1.4pt}

%% file: tables/ablation_for_guidance_scale.tex
\setlength{\tabcolsep}{4pt}
\vspace{0cm}
\begin{table}[h]
\begin{center}
\caption{Ablation study on shape embedding guidance scale $s_s$ and image embedding guidance scale $s_p$. The values are [FID/ CLIP similarity]. }
\label{table:ablation_for_guidance_scale}
\begin{tabular}{l c c c c}
\hline\noalign{\smallskip}
\diagbox[dir=NW]{$s_p$}{$s_s$} & 1.0 & 3.0 & 5.0 & 10.0\\
\noalign{\smallskip}
\hline
\noalign{\smallskip}
1.0 & 65.18/0.75934 & 61.20/0.76435& 57.05/0.76149& 58.80/0.75882\\
3.0 & 55.09/0.77800& 55.18/0.77538 & 53.60/\textbf{0.77803} & 53.30/0.77494\\
5.0 & \textbf{53.21}/0.77642& 55.00/0.77524& 53.43/0.77683& 53.86/0.77343\\
10.0 & 54.82/0.77611& 54.82/0.77543& 54.63/0.77643& 54.91/0.77689\\
\hline
\end{tabular}
\end{center}
\vspace{0cm}
\end{table}
\setlength{\tabcolsep}{1.4pt}

%% file: aticle_architectures/conclusion.tex
\section{Conclusion}
This paper proposes a two-stage diffusion model for 3D generation from a single image, that was trained on a highly compressed latent space. To obtain a compressed latent space, we add learnable parameters in the projecting process from 3D to 2D, and we use 3D-aware cross-attention to further enhance the latent. Instead of generating latent conditioned solely on image embedding, we additionally condition on the shape embedding predicted by the diffusion prior model. Compress3D achieves high-quality generation results with minimal training data and training time, showcasing its versatility and adaptability across diverse scenarios.